%% file: sample.tex
\documentclass[sigconf]{aamas}

\usepackage{hyperref}

\usepackage{balance} % for balancing columns on the final page
\usepackage{hyperref}
\usepackage{url}
 \usepackage{graphicx}
\usepackage{wrapfig}
\usepackage{algorithm}
\usepackage{algpseudocode}
\usepackage{amsmath}

\usepackage{algpseudocode}
\usepackage{wrapfig}
\usepackage{amsmath}
\usepackage{geometry}
\usepackage{booktabs}
\usepackage{multirow}
\usepackage{graphicx}
\theoremstyle{definition}
\newtheorem{definition}{Definition}
\theoremstyle{theorem}
\newtheorem{theorem}{Theorem}

\usepackage{booktabs} % For better table formatting
\usepackage{multirow}
\usepackage{adjustbox}
\usepackage{colortbl} % For cell coloring
\usepackage{xcolor}   % For color definition
\usepackage{array, enumitem} % Add this in the preamble for custom formatting

\usepackage{titletoc}
\usepackage{tocloft}
\usepackage{booktabs} % Add this to your preamble for nicer table lines
\usepackage{adjustbox} % Add this to your preamble if you want to adjust the table width
\setcopyright{ifaamas}
\acmConference[AAMAS '25]{Proc.\@ of the 24th International Conference
on Autonomous Agents and Multiagent Systems (AAMAS 2025)}{May 19 -- 23, 2025}
{Detroit, Michigan, USA}{A.~El~Fallah~Seghrouchni, Y.~Vorobeychik, S.~Das, A.~Nowe (eds.)}
\copyrightyear{2025}
\acmYear{2025}
\acmDOI{}
\acmPrice{}
\acmISBN{}

\acmSubmissionID{111}

\title[AdaCred]{AdaCred: Adaptive Causal Decision Transformers with Feature Crediting}

\author{Hemant Kumawat}
\affiliation{
  \institution{Georgia Institute of Technology}
  \city{Atlanta}
  \country{USA}}
\email{hkumawat6@gatech.edu}

\author{Saibal Mukhopadhyay}
\affiliation{
  \institution{Georgia Institute of Technology}
  \city{Atlanta}
  \country{USA}}
\email{saibal.mukhopadhyay@ece.gatech.edu}

\begin{abstract}
Reinforcement learning (RL) can be formulated as a sequence modeling problem, where models predict future actions based on historical state-action-reward sequences. Current approaches typically require long trajectory sequences to model the environment in offline RL settings. However, these models tend to over-rely on memorizing long-term representations, which impairs their ability to effectively attribute importance to trajectories and learned representations based on task-specific relevance. In this work, we introduce AdaCred, a novel approach that represents trajectories as causal graphs built from short-term action-reward-state sequences. Our model adaptively learns control policy by crediting and pruning low-importance representations, retaining only those most relevant for the downstream task. Our experiments demonstrate that AdaCred-based policies require shorter trajectory sequences and consistently outperform conventional methods in both offline reinforcement learning and imitation learning environments.
\end{abstract}


\newcommand{\BibTeX}{\rm B\kern-.05em{\sc i\kern-.025em b}\kern-.08em\TeX}

\begin{document}

%%% The following commands remove the headers in your paper. For final 
%%% papers, these will be inserted during the pagination process.

\pagestyle{fancy}
\fancyhead{}

%%% The next command prints the information defined in the preamble.

\maketitle 

%%%%%%%%%%%%%%%%%%%%%%%%%%%%%%%%%%%%%%%%%%%%%%%%%%%%%%%%%%%%%%%%%%%%%%%%

\section{Introduction}
% Recent advancements in Reinforcement Learning (RL) have drawn significant inspiration from the successes of sequence modeling, particularly Transformer architectures. Traditionally, RL problems \cite{kumawat2024robokoop}\cite{sutton2018reinforcement} were often represented through Markov Decision Processes (MDPs), which simplify temporal dependencies by assuming that the current state encapsulates all necessary historical information. However, such formulations can struggle with long-term dependencies and credit assignment, which are essential in many challenging real-world applications. Emerging research has explored the potential of transforming offline RL 
%  \cite{levine2020offlinereinforcementlearningtutorial} into a sequence modeling task 
%  \cite{chen2021decision, offline-rl-lev}, leveraging architectures like Transformers 
%  \cite{vaswani2017attention} to better capture the complexities inherent in state-action-reward sequences. By treating trajectories as generative models, these approaches enable improved policy optimization in both online and offline RL settings.

 Transformer-based approaches, such as Decision Transformer (DT) \cite{chen2021decision} and its variants \cite{wu2024elastic, janner2021offline, shang2022starformer}, have redefined offline reinforcement learning (RL) as a supervised learning problem by leveraging extended historical contexts to relax the strict Markov assumption. These models use self-attention mechanisms to process entire trajectory sequences, enabling the capture of dependencies across long temporal horizons. Despite their advantages, these methods face three primary challenges: (a) \textbf{\textit{Long Sequence Memory:}} The requirement for long trajectory sequences to predict individual actions, (b) \textbf{\textit{Suboptimal Trajectories:}} Limited adaptability in processing datasets containing suboptimal trajectories, and (c) \textbf{\textit{Credit Assignment:}} Lack of an explicit mechanism to assess and weigh the importance of trajectories and representations according to task-specific relevance.
These challenges arise primarily from the inherent limitations of Transformers in RL, which, while proficient at memorizing sequences, struggle to assign appropriate importance to task-relevant trajectories \cite{ni2023when}. Consequently, DT and its variants often require extended sequences---spanning 30 to 50 past trajectories---just to generate a single action. Moreover, these methods exhibit poor adaptability in scenarios involving suboptimal trajectories. By attending to entire sequences indiscriminately, they risk learning policies based on suboptimal behavior, thereby limiting overall performance.
Recent advancements, such as Elastic Decision Transformer \cite{wu2024elastic}, aim to address these shortcomings by selectively "stitching" optimal segments from suboptimal trajectories. These methods estimate the expected reward for each segment and selectively attend to recent observations when the estimated reward is low. However, the exclusion of earlier observations when a trajectory is deemed suboptimal risks diluting essential relational information or overemphasizing redundant recent data points, which can ultimately hinder performance.

% \begin{wrapfigure}{r}{3cm}
%     \centering
%     \includegraphics[width=\linewidth]{figures/causal.png}
%     \caption{Features representation as causal graph}
%     \label{fig:causal}
% \end{wrapfigure} 

\begin{figure}
    \centering
    \includegraphics[width = 0.8 \linewidth]{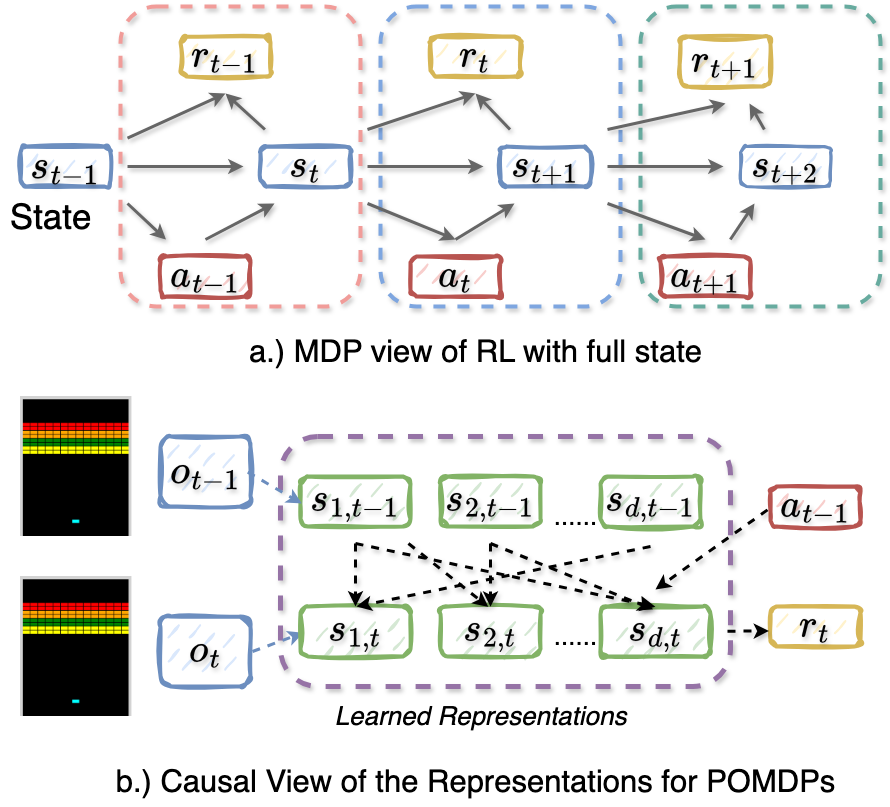}
    \caption{POMDP view of RL}
    \label{fig:causal_1}
\end{figure}

\begin{figure}
    \centering
    \includegraphics[width=\linewidth]{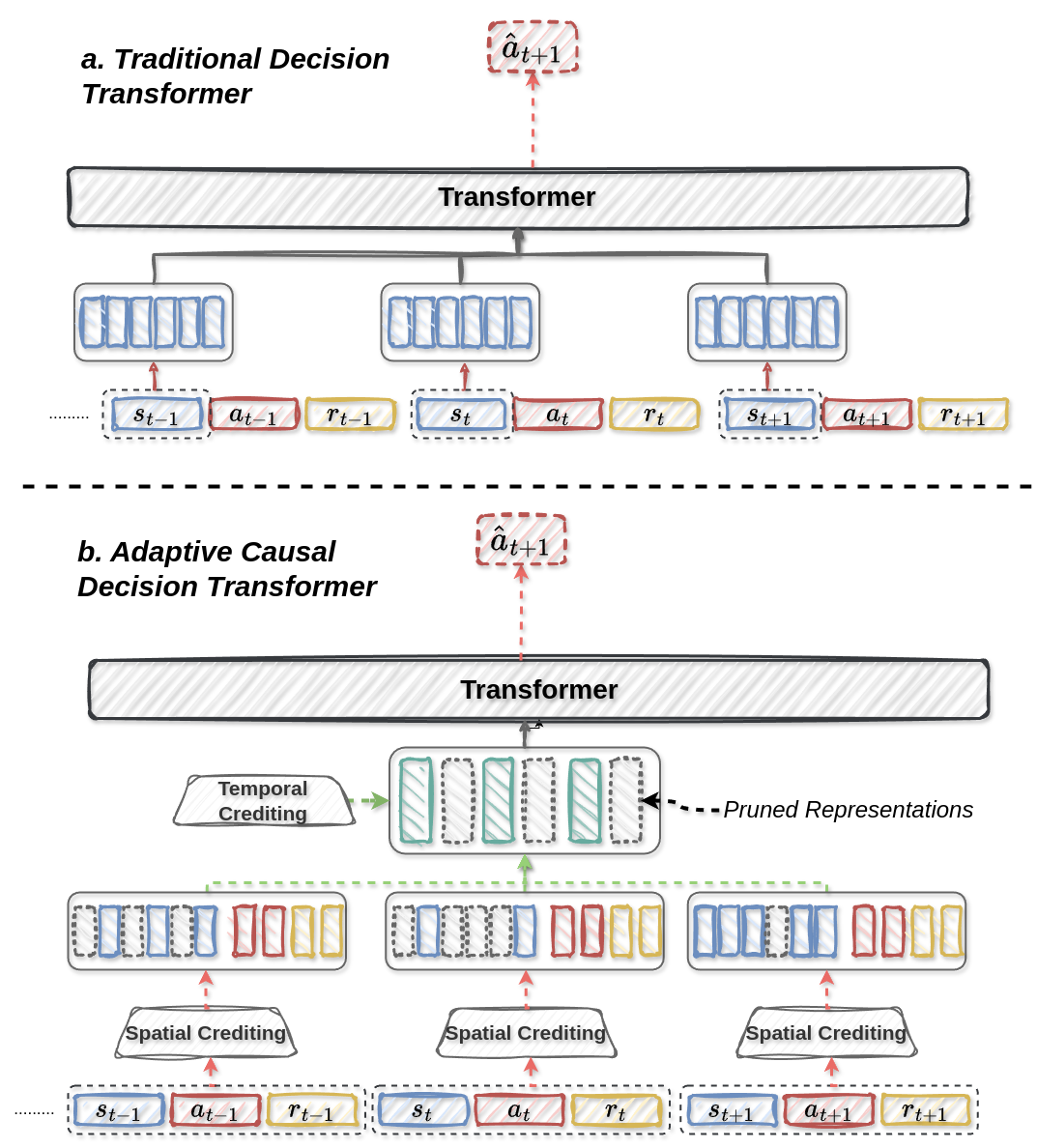}
    \caption{Our model design}
    \label{fig:model1}
\end{figure}

Unlike prior works that focus on learning and retaining all representations, we recast the problem as learning a \textbf{parsimonious causal graphical representation}. Each representation learned from the observations is treated as a node in a causal graph, and we reformulate the original task as identifying and pruning causal variables that are redundant for policy learning, based on the dynamically evolving causal structure. This adaptive approach allows the model to capture essential dependencies while discarding irrelevant data. Figure \ref{fig:causal_1}a illustrates the MDP perspective of reinforcement learning when the full state is available, whereas Figure \ref{fig:causal_1}b depicts the causal structure of representations when the full state is not accessible. This representation explicitly encodes the structural relationships among different observations across the temporal domain.  \textit{The core hypothesis of our approach is that, by structuring the learned representations as a causal graph, not all representations contribute to the future reward. Consequently, {low-credit representations}—those that do not influence future rewards—can be pruned without negatively impacting the agent's performance. This pruning mechanism selectively filters out irrelevant state components, allowing the model to focus on the most {informative parts} of the trajectory.} Our approach significantly reduces the \textbf{effective memory length} of the sequence, allowing the model to allocate computational resources more efficiently by prioritizing high-credit representations. By selectively pruning low-credit representations, our framework not only mitigates the negative impact of suboptimal trajectories but also reinforces critical relationships within the sequence, thus enhancing decision-making quality. Figure \ref{fig:model1} illustrates the overall design of spatial and temporal credit assignment for decision transformer models.

We support our framework with rigorous theoretical results, where we prove that the proposed {credit assignment mechanism} leads to the identification of a minimal set of latent state components that are sufficient for effective policy learning. We formally demonstrate that the pruning process does not negatively impact the agent’s ability to maximize future rewards, as irrelevant features are effectively removed while preserving the necessary structural relationships in the causal graph. We empirically validate our method on well-established benchmarks, including the {Atari} and {Gym environments}, where our model demonstrates superior decision-making and sequence generation capabilities compared to Decision Transformer and its variants. The experimental results reveal that our {credit assignment and pruning mechanisms} lead to more effective policy learning while maintaining computational efficiency. Our method introduces minimal overhead during training, yet yields significant improvements in performance. By surpassing state-of-the-art methods in both offline and imitation learning tasks, our approach opens new avenues for future research in reinforcement learning models that emphasize structured and efficient representation learning.

In this paper, we make the following key contributions: 

\begin{itemize}
    \item  We introduce \textbf{AdaCred}: Adaptive Causal Decision Transformers, a novel approach that addresses the challenge of suboptimal sequence representations in Decision Transformer and related models by leveraging causal latent graph structures with feature crediting.
    \item The causal latent graph representation allows AdaCred to assign credit to individual sequence representations and prune low-importance components, identifying optimal sequences even from suboptimal trajectories, thereby improving decision-making efficiency.
    \item We provide a rigorous theoretical framework for our model and our experimental results on standard benchmarks demonstrate the superior performance of AdaCred in both offline and imitation learning settings.
\end{itemize}

\section{Related Work}

\paragraph{Transformer-based RL Methods}
One of the most notable Transformer based models for offline RL is the {Decision Transformer (DT)} \cite{chen2021decision}, which recasts RL as a supervised learning task, treating trajectory sequences as input and predicting the optimal actions. The DT and its variants \cite{wu2024elastic, janner2021offline, shang2022starformer} have demonstrated strong performance across a range of benchmarks, thanks to their ability to model long-term dependencies via self-attention mechanisms. However, these methods face challenges when applied to suboptimal trajectories or when dealing with long sequences, often requiring the model to attend to up to 50 past trajectories to generate a single action. Recent efforts, such as {Elastic Decision Transformer (EDT)} \cite{wu2024elastic}, address these limitations by selectively attending to optimal trajectory segments and dynamically stitching together segments with high estimated rewards. However, such approaches may lose essential information by prematurely discarding earlier observations, particularly in long-horizon tasks. Our approach extends the Transformer-based RL framework by introducing \textbf{adaptive credit assignment} to mitigate the reliance on long sequences and suboptimal data, allowing the model to focus on task-relevant representations.

\textit{Reinforcement Learning with Suboptimal Data}
Incorporating suboptimal trajectories into RL models has proven to be a challenging task, as suboptimal data can significantly degrade the quality of learned policies. Methods such as {Conservative Q-Learning (CQL)} \cite{kumar2020conservative} have attempted to mitigate the effects of suboptimal data by introducing conservative constraints during policy optimization. Similarly, {Data-Driven Offline RL} \cite{levine2020offlinereinforcementlearningtutorial} aims to learn policies from imperfect datasets while minimizing the impact of suboptimal actions and states. Recent works have also explored {sequence stitching} techniques \cite{wu2024elastic} to combine high-reward segments from suboptimal trajectories, improving overall performance. However, these methods often fail to adapt dynamically to the changing relevance of different parts of the sequence. 

\section{Adaptive Causal Decision Transformer}

\subsection{Theoretical Discussion}
In this section, we present a formal theoretical framework that underpins our AdaCred model. The motivation for this analysis is to rigorously define the representation of latent states in reinforcement learning, introduce a feature credit assignment mechanism, and study the effects of pruning latent representations on policy learning.

\paragraph{POMDPs in Offline Reinforcement Learning} We consider a partially observable Markov decision process (POMDP) 
$M=(O, A, P,$ $ R, \gamma, T)$, where at each time step \( t \in \{1, \dots, T\} \), an agent observes \( o_t \in O \), selects an action \( a_t \in A \) based on the history of past observations and actions, \( h_{1:t} := (o_{1:t}, a_{1:t-1}) \in H_t \), and receives a reward \( r_t \). The reward is distributed as \( r_t \sim R_t(h_{1:t}, a_t) \), and the next observation \( o_{t+1} \) is sampled from the transition dynamics \( P(\cdot | h_{1:t}, a_t) \). The initial observation \( o_1 \) is sampled from the distribution \( P(o_1) \). The problem is characterized by a time horizon \( T \in \mathbb{N}^+ \cup \{+\infty\} \) and a discount factor \( \gamma \in [0, 1] \). In the offline setting, the agent has access to a trajectory \( \tau = \{o_1, a_1, r_1, \dots, o_t, a_t, r_t\} \), which reflects the agent's experience of the environment. For sequence modeling in offline RL, we employ Transformer architectures, treating observations, actions, and rewards as input tokens to approximate the optimal action \( P(a_t | o_{1:t}, a_{1:t-1}, r_{1:t-1}) \).

\paragraph{Graphical Representation of Latent State Transitions}
We denote the underlying latent states for \( o_t \) as \( \mathbf{g}_t = (g_{1,t}, \dots, g_{d,t})^\top \), where \( d \) is the dimensionality of the latent states. The environment’s generative process can be described by the following transition, observation, and reward functions for each dimension of \( \mathbf{g} \):

\[
\begin{aligned}
    g_{i,t} &= f_i(c^{g \to g}_i \odot \mathbf{g}_{t-1}, c^{a}_i \cdot \mathbf{a}_{t-1}, c^{r}_i \cdot \mathbf{r}_{t-1}, \epsilon^{g}_{i,t}), \quad \text{for} \ i = 1, \dots, d, \\
    o_t &= g(c^{g \to o} \odot \mathbf{g}_t, \epsilon^{o}_t), \\
    r_t &= h(c^{g \to r }_r \odot \mathbf{g}_{t-1}, c^{a \to r} \cdot \mathbf{a}_{t-1}, \epsilon^{r}_t),
\end{aligned}
\]

where \( \odot \) represents element-wise multiplication, and \( \epsilon^{g}_t, \epsilon^{o}_t, \epsilon^{r}_t \) are independent noise terms. The terms \( c^{. \to .} \) are binary masks indicating causal relationships between variables. The latent states \( \mathbf{g}_{t+1} \) form a Markov decision process (MDP): given \( \mathbf{g}_t \) and \( \mathbf{a}_t \), the next latent state \( \mathbf{g}_{t+1} \) is independent of previous states and actions. The observed signals \( o_t \) are generated from the latent states \( \mathbf{g}_t \), and rewards are determined by both the latent states and actions.

\begin{definition}[\textbf{{Graphical Representation of Latent State Transitions}}]
Given a causal graph \( G = (V, E) \), where the nodes \( V \) represent the latent state components \( \mathbf{g}_t = (g_{1,t}, \dots, g_{d,t})^\top \), and edges \( E \subseteq V \times V \) represent causal relationships between them, a binary mask \( \mathbf{c}^{g \to g} \in \{0,1\}^{d \times d} \) encodes the presence of edges. The \( i,j \)-th entry of \( \mathbf{c}^{g \to g} \) is 1 if \( g_{i,t+1} \) is influenced by \( g_{j,t} \), and 0 otherwise. Similarly, binary masks \( \mathbf{c}^{a \to g}, \mathbf{c}^{r \to g} \in \{0,1\}^{d} \) represent causal effects of actions and rewards on latent states.
\end{definition}

% \subsection{Feature Credit Assignment}
For each latent state component \( g_{i,t} \), the crediting mechanism outputs a binary mask \( c_{i,t} \in \{0,1\} \), where \( c_{i,t} = 1 \) if \( g_{i,t} \) contributes to the future reward \( r_{t+k} \) for some \( k > 0 \), and 0 otherwise. The binary mask is learned through a credit assignment process that leverages causal relationships encoded in \( \mathbf{c} \), as well as the temporal dependence of state features on rewards. We now present key theoretical results, starting with the minimal sufficient representations required for policy learning.

\begin{theorem}[\textbf{Minimal Sufficient Representations for Policy Learning}]
Under the assumption that the causal graph \( G \) is Markov and faithful to the observed data, the set of minimal latent states $\mathbf{g}_t^{\min} \subseteq \mathbf{g}_t$ is defined as:
$g_{i,t} \in \mathbf{g}_t^{\min}$   if  $c_{i}^{g \to r} = 1$ { or } $g_{i,t}$ has a directed path to a future reward through other latent states
This set forms a minimal and sufficient representation for policy learning, ensuring that only these latent state components are necessary to maximize the expected future rewards. (For full proof, please refer to \textcolor{blue}{Appendix B.1})
\end{theorem}

\begin{proof}[Proof Sketch]
The proof follows from the Markov property, which ensures that future rewards depend only on the current latent states and actions. By d-separation, any state \( g_{i,t} \) that is not part of \( \mathbf{g}_t^{\min} \) is conditionally independent of future rewards given \( \mathbf{g}_t^{\min} \), and thus, pruning these states does not affect the policy's ability to predict future rewards. The remaining states in \( \mathbf{g}_t^{\min} \) are those that directly or indirectly contribute to future rewards, making them both necessary and sufficient for policy learning.
\end{proof}

\begin{theorem}[\textbf{Structural Identifiability}]
Suppose the underlying latent states \( \mathbf{g}_t \) are observed, i.e., the environment follows a Markov Decision Process (MDP). Then, under the Markov condition and faithfulness assumption, the structural matrices \( \mathbf{c}^{g \to g}, \mathbf{c}^{a \to g}, \mathbf{c}^{g \to r}, \mathbf{c}^{a \to r} \) are identifiable.  (For full proof, please refer to \textcolor{blue}{Appendix B.2})
\end{theorem}

\begin{proof}[Proof Sketch]
The identifiability of the structural matrices follows from conditional independence relations in the data. By the faithfulness assumption, the observed conditional independencies reflect the structure of the causal graph \( G \). Since the dynamic Bayesian network (DBN) is time-invariant and adheres to the Markov property, the causal dependencies encoded in the matrices \( \mathbf{c} \) can be identified by analyzing the conditional distributions of the latent states, actions, and rewards. This guarantees the identifiability of the structural matrices.
\end{proof}

\textit{\textbf{Causal Graph Pruning}} 
As previously discussed, not all latent states contribute directly to future actions and rewards. During learning, we can identify minimal sufficient state representations that are essential for task performance. Based on the theoretical results above, we can classify the state representations and prune the causal graph accordingly. This pruning process aims to reduce redundancy in the latent space, thus improving both the efficiency and generalization capability of the model.

Following the structural analysis of latent states, we define two classes of state representations, adapted from \cite{zhang2011causal, yang2024towards}:

1. \textbf{Compact State Representation (\( \mathbf{g}_{t}^{c} \))}: This class includes latent state variables that either influence the observation \( o_t \), contribute to the reward \( r_{t+1} \), or affect other latent states \( g_{j,t+1} \) at the next time step. Formally, a state variable \( g_{i,t} \) is part of this compact set if any of the following structural dependencies hold:
\(
   c^{g \to o} = 1 \quad \text{or} \quad c^{g \to r} = 1 \quad \text{or} \quad c^{g \to g_j} = 1, \quad \forall j \neq i.
\)
These variables form the core of the latent state space that is essential for policy learning and prediction of future rewards.

2. \textbf{Non-Compact State Representation (\( \mathbf{g}_{t}^{\bar{c}} \))}: This class contains state variables that do not satisfy the criteria for compact representations. These variables can be pruned without loss of policy learning capability, as they neither influence the immediate observations nor the future rewards.

By distinguishing between compact and non-compact state representations, we can simplify the causal graph by removing non-essential latent states. To further facilitate the pruning process, we introduce a regularization term that promotes sparsity in the structural masks \( c^{. \to .} \) by encouraging non-essential dependencies to be pruned. Inspired by the edge-minimality principle \cite{zhang2011causal}, we define the regularization term as follows:
\[
J_{\text{reg}} = -\lambda_{\text{reg}} \left( \| c^{g \to o} \|_1 + \| c^{g \to r} \|_1 + \| c^{g \to g} \|_1 + \| c^{a \to g} \|_1 + \| \theta \|_1 \right),
\]
where \( \lambda_{\text{reg}} \) is a regularization coefficient, and \( \| \cdot \|_1 \) represents the \( \ell_1 \)-norm, which encourages sparsity. Incorporating this regularization term into the objective function \( J \) enables simultaneous pruning and model training. The presence of this regularization term naturally leads to the selective removal of non-compact state variables, as entries in \( c^{. \to .} \) transition from 1 to 0, promoting a sparse, efficient representation of the causal relationships. By reducing the number of latent states and their interactions, the model becomes more interpretable and generalizes better to new tasks.

% \begin{wrapfigure}{r}{4cm}
%     \centering
%     \includegraphics[width=\linewidth]{figures/spatial.png}
%     \caption{Spatial Transformer with Crediting. Output of spatial transformer is sent to the next spatial transformer}
%     \label{fig:spatial_trans}
% \end{wrapfigure} 

\subsection{AdaCred Algorithm}

\begin{figure}
    \centering
    \includegraphics[width=\linewidth]{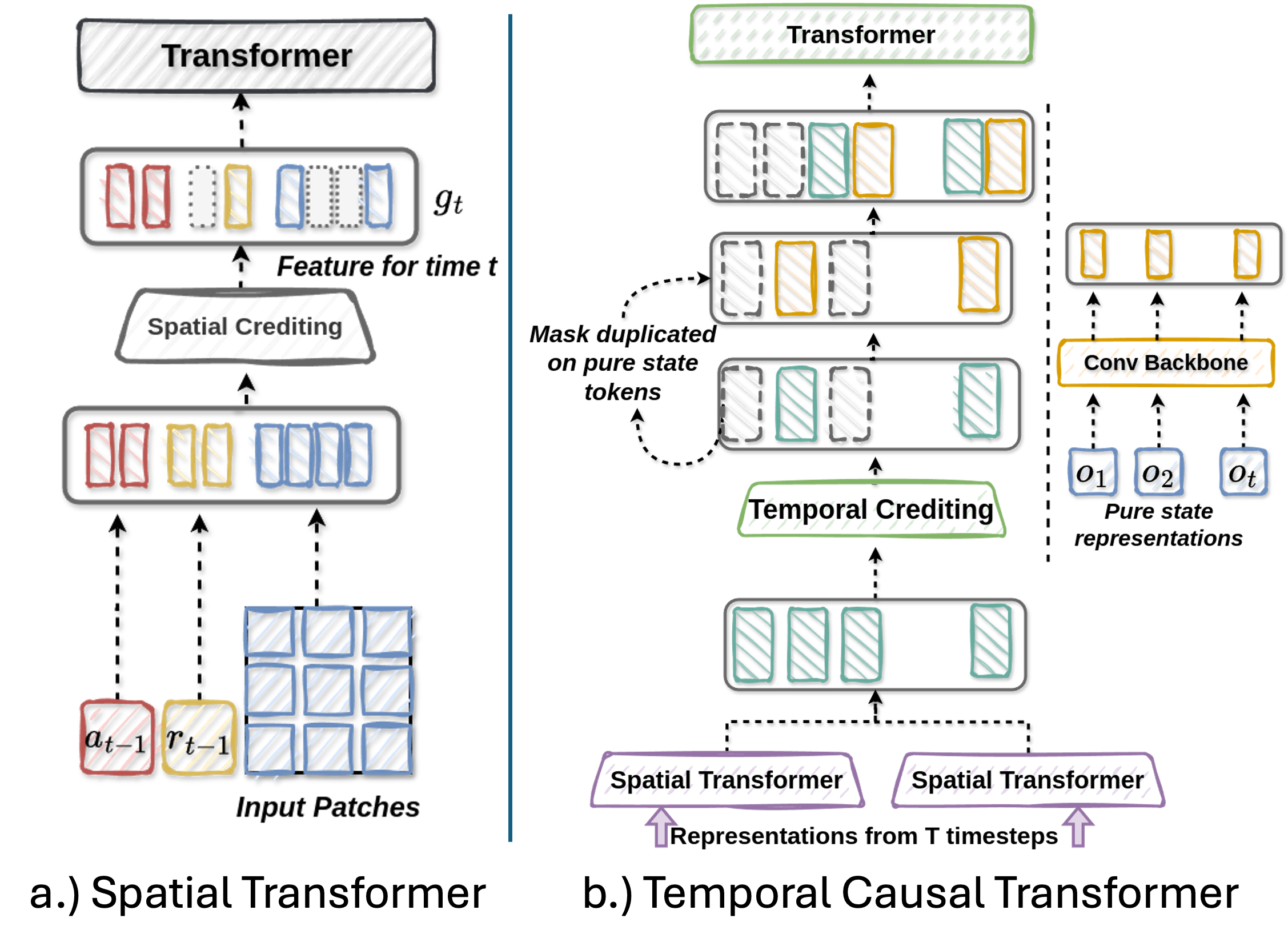}
    \caption{a.) Spatial Transformer with Crediting. Output of spatial transformer is sent to the next spatial transformer. b.) Temporal Causal Transformer with Crediting }
    \label{fig:spa_temp}
\end{figure}

To effectively separate the learning of spatial and temporal representations, we design our causal decision transformer with two distinct components: (a) a Spatial Transformer and (b) a Causal Temporal Transformer. The Spatial Transformer learns spatial representations from the state, action, and reward at each individual time step by performing self-attention on state-action-reward tokens within a single time-step window. These spatial representations are then passed to the Causal Temporal Transformer, which combines them with the pure state representations to model the entire sequence of states. The final action predictions are generated by a prediction decoder. We address the challenge of temporal and spatial feature credit assignment by introducing a feature crediting mechanism that quantifies the importance of latent representations over time.  At each time step \( t \), the agent observes a partial state and produces a latent representation \( g_t \in \mathbb{R}^d \). We introduce a crediting function \( \sigma: \mathbb{R}^d \to [0, 1]^d \) to compute token selection probabilities for each latent feature, which represent the probability that a given feature contributes significantly to future rewards. 
The feature crediting mechanism outputs a binary mask, \( m_t \in \{0, 1\}^d \), . The resultant mask \( m_t \) serves as a gating mechanism, assigning high values to those features deemed influential in maximizing future rewards and low values to those with minimal impact.  Each of the spatial and causal temporal transformer block has this crediting function. In the following sections, we will discuss the design and implementation of these components in detail.

{\textit{Spatial Transformer}} We adopt an Observation-Action-Reward (OAR) grouping mechanism, following a similar approach as in \cite{shang2022starformer}. Each trajectory is partitioned into segments comprising the previous action, the corresponding reward, and the current observation, highlighting the causal interrelationships among these elements. To obtain a fine-grained representation of the state, we partition each input state into non-overlapping spatial patches, thereby enabling localized interactions between actions, rewards, and regions of the state. The action and reward are linearly embedded to produce corresponding representations, which are then combined with the observation token embeddings. This results in an initial representation set of observation, action, and reward tokens, defined as:\(
S_0^t = \{s_{a_{t-1}}, s_{r_{t}}, s_{o_{1, t}}, s_{o_{2, t}}, \ldots, s_{o_{n, t}} \},
\) where \( s_{a_{t-1}} \), \( s_{r_{t}} \), and \( s_{o_{i, t}} \) represent the embeddings for actions, rewards, and observation patches, respectively. This is shown in figure \ref{fig:spa_temp}a where spatial crediting function adaptively computes the tokens to be processed by next layer and prunes the low credit ones .

% \begin{wrapfigure}{r}{0.6\linewidth}
%     \centering
%     \includegraphics[width=\linewidth]{figures/temporal_2.png}
%     \caption{Temporal Causal Transformer with Crediting  }
%     \label{fig:temporal}
% \end{wrapfigure} 
\begin{figure*}
    \centering
    \includegraphics[width=\linewidth]{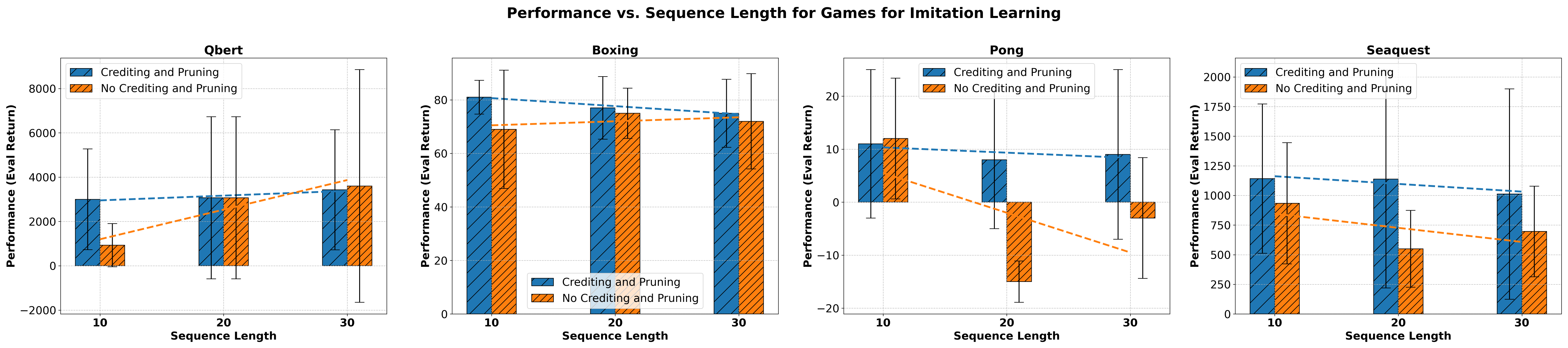}
    \caption{Performance for 75\% Spatial and Temporal Crediting}
    \label{fig:imitation_atari}
\end{figure*}

To process these tokens, we employ a Transformer-based architecture. Specifically, the Spatial Transformer layer maps token representations from the previous layer (\(S_{t}^{l-1}\)) to the current layer (\(S_{t}^l\)) using the transformation:\(
S_{t}^l = F_{\text{step}}^l(S_{t}^{\text{selected}, l-1}),
\) where \( S_{t}^{\text{selected}, l-1} \) represents the subset of features selected by a spatial token policy, which determines the relevant spatial tokens to be processed by the Spatial Transformer block. This token policy serves as a crediting mechanism, selectively pruning unnecessary spatial features before passing them to the subsequent layer, thereby enhancing computational efficiency. For each Spatial Transformer layer \( l \), an aggregated Observation-Action-Reward representation \( g_t^l \in \mathbb{R}^D \) is generated by pooling the output tokens (\( S_{t}^l \in \mathbb{R}^{n \times d} \)), computed as: \(
g_t^l = \text{FC}([\mathrm{concat}(S_{t}^l)]) + e_{t}^{\text{temporal}},
\) where \( \mathrm{concat}(\cdot) \) denotes the concatenation of the tokens within each group, and \( e_{t}^{\text{temporal}} \in \mathbb{R}^D \) is a temporal positional embedding. The resulting OAR representation \( g_t^l \) is then passed to a Temporal Causal Transformer layer for long-term temporal modeling, effectively capturing both local and global dependencies across the trajectory.

% \begin{figure}
%     \centering
%     \includegraphics[width=\linewidth]{figures/temporal_2.png}
%     \caption{Temporal Causal Transformer with Crediting  }
%     \label{fig:temporal}
% \end{figure}

{\textit{Temporal Causal Transformer:}} To effectively model long-term sequences, we introduce the Temporal Causal Transformer, which utilizes the learned spatial representations from the Spatial Transformer, along with pure state representations embedded without any explicit action or review, processed by an encoder.

The pure state representations are defined as follows: $h^0_t = \text{Conv}(o_t) + e^{\text{temporal}}_t$, where $o_t$ represents the image input at time $t$. The output of the Spatial Transformer and the learned convolutional state embedding are combined into a sequence, preserving the temporal order in which they were received: $Y^l_{\text{in}} = \{g^l_1, h^{l-1}_1, g^l_2, h^{l-1}_2, \dots, g^l_T, h^{l-1}_T\}$.  This Transformer layer takes input from each intermediate spatial layer, rather than only the final spatial representation after all layers. Similar to the Spatial Transformer, we employ a token policy that removes certain tokens based on the token policy. The intermediate output of the Temporal Causal Transformer is given by: $Y^l_{\text{out}} = F^l_{\text{sequence}}(Y^l_{\text{in}})$ where similar to the spatial transformer the credit function determines which representations to prune adaptively. This is shown in the figure \ref{fig:spa_temp}b. From $Y^l_{\text{out}}$, we select tokens at even indices (starting from 1) to be the pure state tokens $h^l_i := y^l_{\text{out}; 2i}$. These tokens are then passed to the next Temporal Causal Transformer layer. The final output of the Temporal Causal Transformer layer is used to predict actions using a linear head, denoted as $\hat{a}$. Algorithm \ref{algo1} shows the overall algorithm for our model for single layer networks.

% \begin{figure}
%     \centering
%     \includegraphics[width=\linewidth]{figures/temporal_2.png}
%     \caption{Caption}
%     \label{fig:enter-label}
% \end{figure}

\textit{\textbf{AdaCred Training Method}} The training process for our Adaptive Causal Decision Transformer framework consists of two main stages: training a static model and co-training the model along with the crediting mechanism, ensuring that both spatial and temporal feature assignments are learned effectively.

\textit{{Stage 1: Static Model Training}}: The initial phase involves training the static components of the AdaCred model, comprising the Spatial Transformer and Temporal Causal Transformer. In this stage, we focus on learning the basic spatial and temporal representations that underpin the entire sequence modeling. The training objective for this stage is to minimize the action prediction loss, which is defined as: \(
\mathcal{L}_{\text{stage1}} = \sum_{t=1}^T \mathcal{L}_{\text{action}}(\hat{a}_t, a_t)
\)
where \(\hat{a}_t\) represents the predicted action at time \(t\), \(a_t\) is the ground-truth action, and \(T\) is the total number of time steps. The action prediction loss, \(\mathcal{L}_{\text{action}}\), measures the discrepancy between the predicted and ground-truth actions.

\textit{{{Stage 2: Policy Network with Feature crediting}}}: In this stage, we focus on jointly training the model with the crediting mechanism, optimizing the binary masks \( m_t \in \{0, 1\}^d \) that govern feature selection for both Spatial and Temporal Causal Transformers. At each time step \( t \), the Spatial Transformer and the Temporal Causal Transformer use a feature crediting mechanism \( \sigma: \mathbb{R}^d \to [0, 1]^d \) to compute token selection probabilities for each latent feature representation \( g_t \). The crediting mechanism assigns importance to each feature, generating binary masks that act as gates, allowing only the most influential features to pass through. During co-training, we optimize both the Spatial Transformer and Temporal Causal Transformer layers to ensure that the selected features (via binary masks) maximize the future rewards while maintaining computational efficiency. The training objective for this stage incorporates both the task prediction loss and an efficiency loss, which accounts for the number of features selected: $\mathcal{L}_{\text{stage2}} = \mathcal{L}_{\text{action}} + \alpha \mathcal{L}_{\text{eff}}$
where \(\mathcal{L}_{\text{action}}\) is the total loss across all action predictions, and \(\mathcal{L}_{\text{eff}}\) measures the efficiency of feature selection. The efficiency loss is calculated as: \(
\mathcal{L}_{\text{eff}} = \text{MSE}\left(\omega_d \cdot \frac{T_{\text{activ}}}{T_{\text{total}}}, \omega_d \cdot \frac{T_{\text{target}}}{T_{\text{total}}}\right)
\) where \(T_{\text{activ}}\) represent the number of activated tokens, while \(T_{\text{target}}\) are the desired activation percentages. The term \(\omega_d\) accounts for the different embedding dimensions at different layers, ensuring that the efficiency loss reflects both the number and the size of the selected features, adequately balancing the computational burden.

\begin{table*}
    \centering
    \begin{adjustbox}{max width=\textwidth}
    \begin{tabular}{lccccccc}
        \toprule
        Dataset & DT & CQL & QR-DQN & REM & Baseline (100\% S \& 100\% T*) & \cellcolor{gray!15} Ours (75\% S \& 75\% T* ) \\
        \midrule
        Assault & 462 ± 139 & 432 & 142 & 350 & 458 ± 166 & \cellcolor{gray!15} \textbf{658 ± 66}  \\
        Boxing & 78.3 ± 4.6 & 56.2 & 14.3 & 12.7 & 78 ± 14 & \cellcolor{gray!15} \textbf{78.4 ± 14.5}\\
        Breakout & 76.9 ± 17.1 & 55.8 & 4.5 & 2.4 & 52 ± 14 & \cellcolor{gray!15} \textbf{77 ± 12} \\
        Pong & 12.8 ± 3.2 & 13.5 & 2.2 & 0.0 & 14 ± 7.68 & \cellcolor{gray!15} \textbf{18 ± 2.54 } \\
        Qbert & 3488 ± 631 &  \textbf{14012} & 0.0 & 0.0 & 1612 ±1860 &\cellcolor{gray!15} 4487 ± 5099 \\
        Seaquest &  \textbf{1131 ± 168} & 685 & 161 & 282 & 890 ± 454 &\cellcolor{gray!15} 968 ± 505 \\
        \bottomrule
    \end{tabular}
    \end{adjustbox}
    \caption{Performance Comparison of Various Methods in Offline RL Setting. (X \% S and Y\% T means X \% of Spatial and Y\% of Temporal tokens have been used for learning the model )}
    \label{tab:performance_comparison_offline_rl}
\end{table*}

% \resizebox{\textwidth}{!}{
% \begin{algorithm}
%     \caption{Feature Crediting and Pruning for Single Spatial and Temporal Transformer}
%     \label{algo1}
%     \begin{algorithmic}[1]
%         \Require Observations $\{o_t\}$, Actions $\{a_t\}$, Rewards $\{r_t\}$, Time Horizon $T$
%         \State Initialize parameters $\theta$ for Transformers
%         \For{each time step $t = 1$ to $T$}
%             \State $g_t \gets \text{SpatialTransformer}(o_t, a_t, r_t; \theta)$
%             \State $\sigma(g_t) \gets \text{CreditFunction}(g_t; \theta)$
%             \State $m_t \gets \text{GumbelSigmoid}(\sigma(g_t))$
%             \State $g_t^{\text{selected}} \gets m_t * g_t$
%             \State $g_t^{\text{processed}} \gets \text{TemporalTransformer}(g_t^{\text{selected}}; \theta)$
%             \State $\sigma_{\text{temporal}}(g_t^{\text{processed}}) \gets \text{CreditFunction}(g_t^{\text{processed}})$
%             \State $m_t^{\text{temporal}} \gets \text{GumbelSigmoid}(\sigma_{\text{temporal}}(g_t^{\text{processed}}))$
%             \State $g_t^{\text{final}} \gets m_t^{\text{temporal}} * g_t^{\text{processed}}$
%             \State $a_t^{\text{pred}} \gets \text{ActionDecoder}(g_t^{\text{final}}; \theta)$
%             \State Observe reward $r_t$ and next observation $o_{t+1}$
%         \EndFor
%         \State \Return Predicted actions $\{a_t^{\text{pred}}\}$, optimized policy $\pi$
%     \end{algorithmic}
% \end{algorithm}
% }

\begin{algorithm}
    \caption{Feature Crediting and Pruning for Single Spatial and Temporal Transformer}
    \label{algo1}
    \begin{enumerate}[left=0pt, label=\arabic*., align=left, labelwidth=0.7cm, labelsep=0.3cm]
        \item \textbf{Input:} Observations $\{o_t\}$, Actions $\{a_t\}$, Rewards $\{r_t\}$, Time Horizon $T$
        \item Initialize parameters $\theta$ for Transformers
        \item \textbf{For} each time step $t = 1$ to $T$:
        \begin{enumerate}[label*=\arabic*., left=.5cm, align=left, labelwidth=1cm, labelsep=0.5cm]
            \item $g_t \gets \text{SpatialTransformer}(o_t, a_t, r_t; \theta)$
            \item $\sigma(g_t) \gets \text{CreditFunction}(g_t; \theta)$
            \item $m_t \gets \text{GumbelSigmoid}(\sigma(g_t))$
            \item $g_t^{\text{selected}} \gets m_t * g_t$
            \item $g_t^{\text{processed}} \gets \text{TemporalTransformer}(g_t^{\text{selected}}; \theta)$
            \item $\sigma_{\text{temporal}}(g_t^{\text{processed}}) \gets \text{CreditFunction}(g_t^{\text{processed}})$
            \item $m_t^{\text{temporal}} \gets \text{GumbelSigmoid}(\sigma_{\text{temporal}}(g_t^{\text{processed}}))$
            \item $g_t^{\text{final}} \gets m_t^{\text{temporal}} * g_t^{\text{processed}}$
            \item $a_t^{\text{pred}} \gets \text{ActionDecoder}(g_t^{\text{final}}; \theta)$
            \item Observe reward $r_t$ and next observation $o_{t+1}$
        \end{enumerate}
        \item \textbf{End For}
        \item \textbf{Output:} Predicted actions $\{a_t^{\text{pred}}\}$, optimized policy $\pi$
    \end{enumerate}
\end{algorithm}

\section{Experiments}

Our experiments aim to address the following key research questions, each corresponding to specific aspects of our study:

\begin{itemize}
    \item \textbf{Performance Comparison}: Does AdaCred outperform Decision Transformers (DT) and its variants in Atari and Gym benchmarks? (See \hyperref[sec1]{\textcolor{blue}{\textit{Section 4.2.1}}})
    \item \textbf{Effectiveness in Feature Pruning}: How effective is AdaCred in identifying and pruning both spatial and temporal features? (See \hyperref[sec2]{\textcolor{blue}{\textit{Section 4.2.2}}})
    \item \textbf{Effect of Pruning percentage on performance}: How much can we prune spatially and temporally in Adacred (See \hyperref[sec3]{\textcolor{blue}{\textit{Section 4.2.3}}})
    \item \textbf{Sparse Task Performance}: How well does AdaCred perform in tasks that require sparse representations? (See \hyperref[sec4]{\textcolor{blue}{\textit{Section 4.2.4}}})
\end{itemize}

\subsection{Experimental setting}

\subsubsection{Dataset} Our experiments encompass both offline reinforcement learning (RL) \cite{offline-rl-lev} and imitation learning via behavior cloning. For offline RL, we use a fixed dataset of sub-optimal trajectory rollouts, which presents challenges such as distribution shift, a key difficulty in offline RL compared to standard online RL settings \cite{kumawat2024robokoop, sutton2018reinforcement}. In the imitation learning scenario, the agent is trained without reward signals or real-time environment interaction, which significantly increases the learning difficulty. Unlike traditional imitation learning, where additional data collection and Inverse Reinforcement Learning are feasible, we work solely with sub-optimal trajectories. Following the approach in \cite{shang2022starformer}, we derive the imitation learning dataset by removing rewards from the offline dataset. We evaluate our model in both discrete and continuous action space environments. For discrete environments, we use the Atari benchmark, selecting six games: Assault, Boxing, Breakout, Pong, Qbert, and Seaquest. We employ 1\% (500,000 transitions) of the DQN replay buffer to facilitate thorough and fair comparisons. For continuous control tasks, we use the D4RL benchmark \cite{fu2020d4rl}, which includes HalfCheetah, Hopper, Walker, and a 2D reacher task with sparse rewards to explore beyond the typical locomotion benchmarks. The datasets for these experiments feature three settings: (1) 'Medium,' which consists of 1 million timesteps generated by a policy with moderate performance (about one-third of an expert); (2) 'Medium-Replay,' which contains the replay buffer data collected during training of the medium policy; and (3) 'Medium-Expert,' combining 1 million timesteps from both medium and expert policies. We report episodic returns as absolute values, averaged over ten random seeds for the Atari games and gym tasks. Each seed is evaluated across ten episodes, with random initial conditions to provide a robust performance assessment. 

% \subsubsection{Dataset} Our experiments cover offline reinforcement learning (RL) \cite{offline-rl-lev} and imitation learning via behavior cloning. For offline RL, we use a fixed dataset of sub-optimal trajectory rollouts, presenting challenges like distribution shift compared to online RL \cite{kumawat2024robokoop, sutton2018reinforcement}. In imitation learning, the agent is trained without rewards or real-time interactions, solely using sub-optimal trajectories. Following \cite{shang2022starformer}, we create the imitation learning dataset by removing rewards from the offline dataset. We evaluate our model in both discrete and continuous action spaces. For discrete tasks, we use six Atari games: Assault, Boxing, Breakout, Pong, Qbert, and Seaquest, using 1\% (500,000 transitions) of the DQN replay buffer. For continuous control, we use the D4RL benchmark \cite{fu2020d4rl} with HalfCheetah, Hopper, Walker, and a 2D reacher task with sparse rewards. The datasets include three settings: 'Medium' (1 million timesteps from a moderately performing policy), 'Medium-Replay' (replay buffer during medium policy training), and 'Medium-Expert' (combination of medium and expert policies). We report episodic returns averaged over ten random seeds, each evaluated across ten episodes with random initial conditions.
\begin{figure*}
    \centering
    \includegraphics[width=\linewidth]{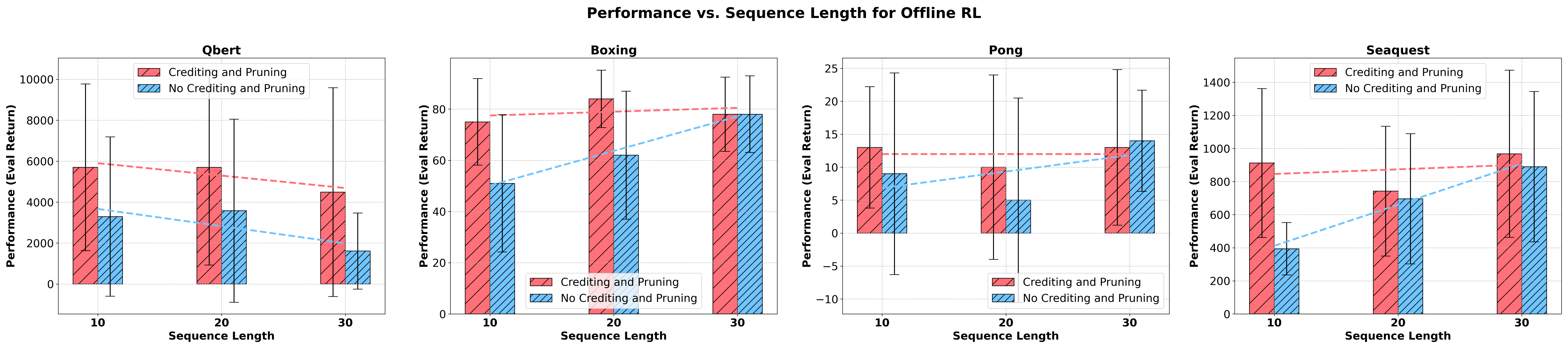}
    \caption{Performance for 75\% Spatial and Temporal Crediting for Offline RL}
    \label{fig:offline_atart}
\end{figure*}

\subsubsection{Baselines}
For baseline comparisons, we include our method using 100\% of spatial and temporal tokens, following the approach in \cite{shang2022starformer}. We evaluate performance by progressively reducing the proportion of spatial and temporal tokens for action prediction. Additionally, we compare against the Decision Transformer \cite{chen2021decision}, its variants with variable-length trajectories \cite{wu2024elastic}, a Q-learning-based method for suboptimal trajectories \cite{qlearning} and EDT \cite{wu2024elastic}. We further benchmark against state-of-the-art non-Transformer offline RL methods, including CQL\cite{kumar2020conservative}, QR-DQN\cite{dabney2018distributional}, REM\cite{agarwal2020optimistic}, and BEAR\cite{kumar2019stabilizing}.

\subsection{Results}

\begin{table*}[ht]
    \centering
    \begin{adjustbox}{max width=\textwidth}
    \begin{tabular}{lcccccccccc}
        \toprule
        Dataset & Environment & DT Sparse & QDT Sparse & BEAR Dense & EDT Dense & AWR & Baseline Dense & \cellcolor{gray!15} Ours Dense & Baseline Sparse & \cellcolor{gray!15} Ours Sparse \\
        \midrule
        Medium-Expert & HalfCheetah & - & - & 53.4 & - & 52.7 & 77.4 ± 1.2 & \cellcolor{gray!15} 77.8 ± 1.3 & 67 ± 1.2 & \cellcolor{gray!15} \textbf{81 ± 1.6} \\
        Medium-Expert & Hopper & - & - & 96.3 & - & 27.1 & 97.9 ± 4.6 & \cellcolor{gray!15} 102.7 ± 5.6 & \textbf{106.2 ± 2.6} & \cellcolor{gray!15} 97.33 ± 3.6 \\
        Medium-Expert & Walker & - & - & 40.1 & - & 53.8 & 108.5 ± 7.6 & \cellcolor{gray!15} 108.8 ± 5.6 & 108 ± 3.6 & \cellcolor{gray!15} \textbf{109 ± 4.6} \\
        Medium & HalfCheetah & 42.2 ± 0.2 & 42.4 ± 0.5 & 41.7 & 42.5 ± 0.9 & 37.4 & 42.4 ± 1.6 & \cellcolor{gray!15} \textbf{42.8 ± 2.6} & 42.1 ± 0.5 & \cellcolor{gray!15} \textbf{42.7 ± 0.7} \\
        Medium & Hopper & \textbf{57.3 ± 2.4} & 50.7 ± 5.0 & 52.1 & 63.5 ± 5.8 & 35.9 & 51.2 ± 2.1 & \cellcolor{gray!15} 55.6 ± 1.2 & 52.11 ± 0.6 & \cellcolor{gray!15} 54.47 ± 2.1 \\
        Medium & Walker & {69.9 ± 2.0} & 63.7 ± 6.4 & 59.1 & 72.8 ± 6.2 & 17.4 & 72.4 ± 1.8 & \cellcolor{gray!15} \textbf{74 ± 3.9} & 38.46 ± 3.8 & \cellcolor{gray!15} \textbf{75.15 ± 4.2} \\
        \bottomrule
    \end{tabular}
    \end{adjustbox}
    \caption{Performance comparison across datasets and environments in Medium and Medium-Expert settings.}
    \label{tab:performance_comparison_sparse}
\end{table*}
\begin{figure}
    \centering
    \includegraphics[width=\linewidth]{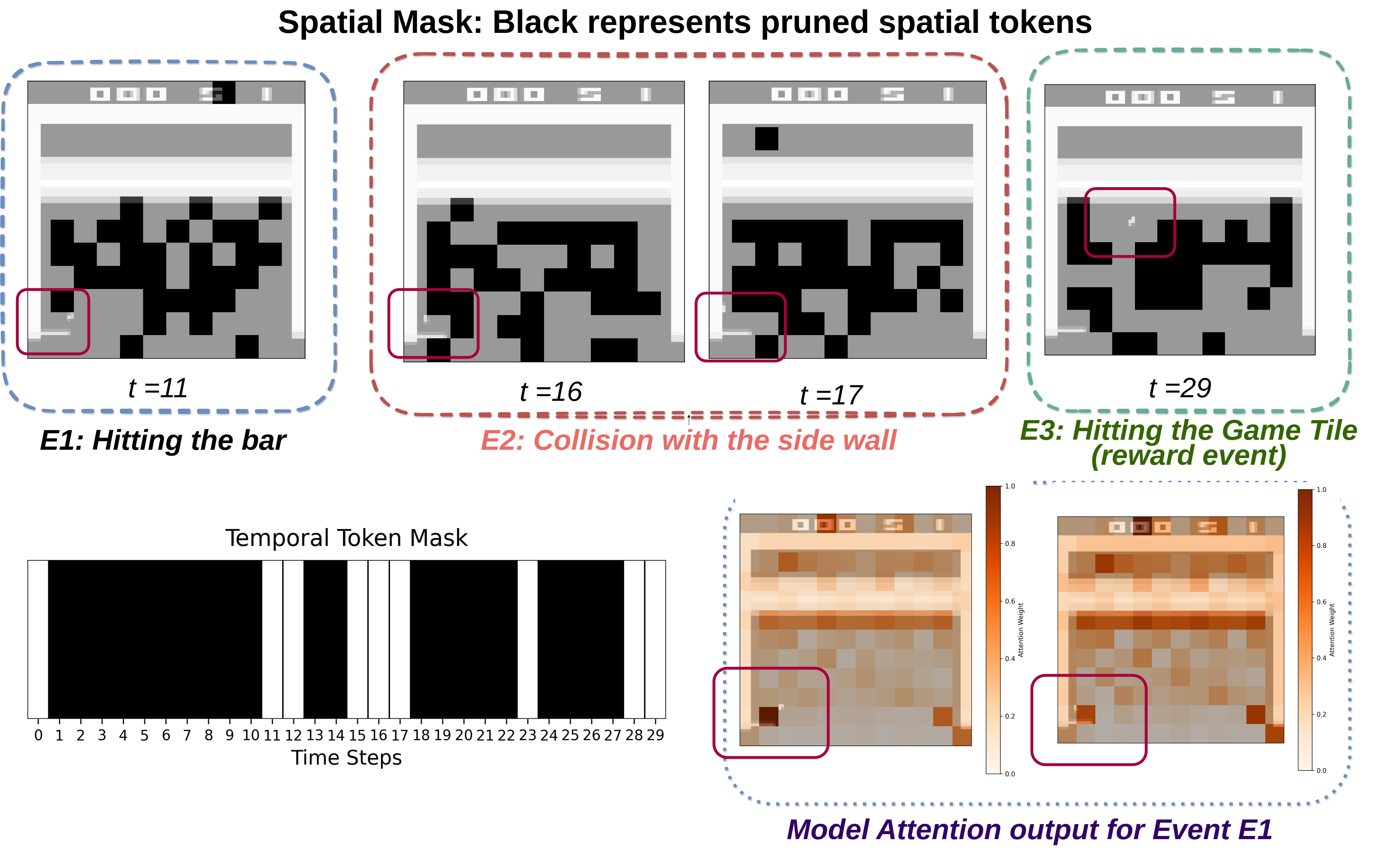}
    \caption{Visualization of our spatial and temporal masks with attention outputs}
    \label{fig:viz}
\end{figure}

\subsubsection{Analysis of Atari Games for Offline Reinforcement Learning and Imitation Learning}
\label{sec1}

We evaluate the effectiveness of our credit assignment and pruning mechanism on Atari games, focusing on both the imitation learning and offline reinforcement learning (RL) settings. To establish a baseline, we first train a model without any credit assignment or pruning mechanisms. We then compare it to a version of the model that incorporates 75\% spatial and temporal pruning. For both models, we use sequence lengths of 10, 20, and 30 steps, consistent with the experimental setup of the Decision Transformer (DT). The results for four Atari environments (Qbert, Boxing, Pong, and Seaquest) are illustrated in Figure \ref{fig:imitation_atari}, while Figure \ref{fig:offline_atart} presents the results for the offline RL setting. In both settings, the baseline model exhibits the highest performance when sequence trajectories of 30 time steps are provided, with a noticeable decline in performance as the sequence length is reduced. This indicates that the baseline model relies on longer sequences to capture relevant information for decision-making. Conversely, our model demonstrates a marked improvement in efficiency, achieving optimal performance with sequences as short as 10 time steps and maintaining consistent performance across all tested sequence lengths. Remarkably, even with just 10 steps, our model consistently outperforms the baseline model's best results with 30-step sequences. This clearly demonstrates that our approach effectively learns to assign credit and prune unnecessary representations, thereby retaining only the most relevant portions of the trajectory data. The model's ability to maintain peak performance with minimal sequence lengths indicates that it learns optimal representations early on in the trajectory, rendering additional steps redundant. This result is particularly significant for resource-constrained environments, as the model's performance remains robust while processing only 33\% of the sequence data compared to the baseline.

In Table \ref{tab:performance_comparison_offline_rl}, we provide a comprehensive performance comparison between our proposed model with 75\% spatial and temporal pruning and various baseline methods, including Decision Transformer (DT), Conservative Q-Learning (CQL), QR-DQN, and REM, across six different Atari games. The metrics presented reflect the average rewards obtained, along with the standard deviation where applicable. Our approach demonstrates superior performance in nearly all tasks, particularly excelling in games that typically require longer-term planning and spatial awareness, such as Assault and Seaquest. For Qbert, which involves complex decision-making and long-term strategy, our model exhibits an impressive performance of 4487 ± 5099, significantly improving upon the baseline (1612 ± 1860) and outperforming DT (3488 ± 631). The results indicate that our model effectively manages credit assignment over longer sequences, even in environments with extensive trajectory dependencies. Overall, Table \ref{tab:performance_comparison_offline_rl} highlights the competitive advantage of our approach across various environments, outperforming or matching traditional methods, including DT and CQL. Notably, while methods like CQL are specifically designed to optimize policies from suboptimal data, they struggle to generalize effectively in these environments, achieving suboptimal rewards. In contrast, our approach, even with 75\% pruning of spatial and temporal tokens, consistently surpasses other models, demonstrating its robustness and capability in both spatial and temporal credit assignment.

\subsubsection{Effectiveness of Spatial and Temporal Crediting} 
\label{sec2}

To demonstrate the model's capability in assigning credit and pruning representations that are not relevant to the task, we present a visualization of the spatial and temporal token masks in Figure \ref{fig:viz}. The figure showcases the Breakout game scenario, where the model was trained for 30 timesteps, highlighting key moments and decision-making processes during gameplay. In the second row of Figure \ref{fig:viz}, the temporal scores from the causal decision transformer are depicted. These scores reflect the significance of different timesteps in relation to action prediction. The sequences numbered 11, 12, 15, 16, 17, 28, and 29 are identified by the model as having the highest importance, suggesting that these time steps contain critical information for credit assignment in decision-making. The top row of Figure \ref{fig:viz} displays the output of the spatial mask from the first layer of the spatial transformer. In this visualization, the black regions correspond to areas of the image that are disregarded by the model, indicating that they provide little or no useful information for the task at hand. From the spatial mask, it is evident that the model learns to focus on the relevant regions containing dynamic elements, such as the paddle and the ball, while ignoring other static parts of the environment. Specifically, the central region of the game screen, which primarily contains unchanging elements, is discarded, as it does not contribute to the decision-making process. The spatial transformer efficiently identifies the regions where important interactions occur, such as the movement of the ball and paddle, which are assigned higher attention scores. This is clearly reflected in the attention maps, where the ball and paddle are shown to receive the highest attention, indicating their critical role in the task. Moreover, the temporal token mask reveals that the model focuses on key events, which can be categorized into three main interactions: a.) \textbf{Collision with the paddle (E1):} The ball's interaction with the paddle is crucial for determining the next trajectory and is therefore heavily weighted. b.) \textbf{Collision with the side wall (E2):} The side wall collision alters the ball's path and is a significant event that affects the gameplay. c.) \textbf{Hitting a game tile (E3):} This event results in a reward, making it an essential moment for credit assignment. The model successfully learns to emphasize these key interactions, which are crucial for effective decision-making. By focusing on these spatial and temporal aspects, the network is able to prune irrelevant representations and learn efficient action policies. This targeted credit assignment mechanism enables our model to outperform others in similar tasks by prioritizing relevant events and discarding extraneous information.

\subsubsection{Effect of pruning percentage on performance}
\label{sec3}

\begin{figure}
    \centering
    \includegraphics[width=\linewidth]{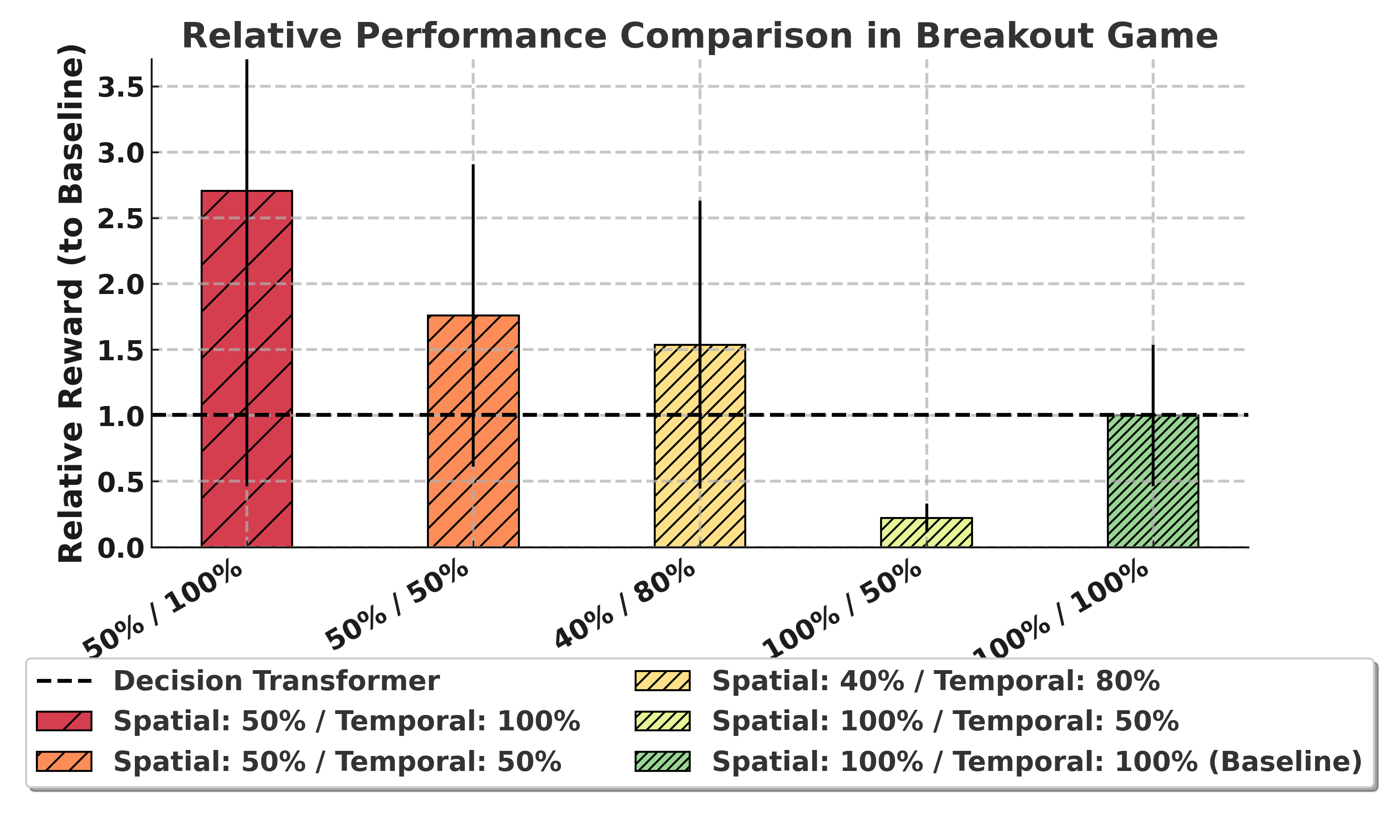}
    \caption{Effect of Spatial and Temporal Pruning}
    \label{fig:pruning_ab}
\end{figure}

In Figure \ref{fig:pruning_ab}, we examine the effect of varying degrees of spatial and temporal pruning on the performance of our model in the Breakout game. The pruning ratios are defined as (X\%, Y\%), where X\% represents the percentage of spatial features retained by the model, and Y\% represents the percentage of temporal features retained. We test the following combinations: (50\%, 50\%), (50\%, 100\%), (40\%, 80\%), and (100\%, 50\%), with the baseline model using 100\% of both spatial and temporal features. The model achieves its highest performance with the (50\%, 100\%) pruning ratio, where 50\% of the spatial features are retained while all temporal features are used. This result supports our hypothesis that, among the 30 sequences analyzed, the majority of redundancies are spatial rather than temporal. In the context of Breakout, the spatial redundancies primarily come from the images, where large portions of the background contain redundant or irrelevant information that does not contribute to decision-making. By pruning these less critical spatial features, the model can generalize better, focusing on the most relevant spatial information, such as the paddle and ball, while retaining the full temporal context to capture important trajectory-based decisions. The combination of spatial pruning and full temporal retention leads to a significant improvement in performance, with the model achieving approximately 2.8 times the baseline performance. On the other hand, the lowest performance is observed with the (100\%, 50\%) pruning ratio, where all spatial features are used but 50\% of the temporal features are pruned. In this case, retaining all spatial features introduces redundant spatial information, which in turn establishes misleading causal relationships in the temporal domain. The absence of sufficient temporal information exacerbates the issue, leading the model to infer incorrect temporal dependencies. If proper temporal pruning had been applied, these erroneous temporal relationships could have been removed, allowing for better decision-making. This pruning ratio significantly underperforms compared to the baseline, emphasizing the critical role of temporal features in tasks requiring long-term planning and time-sensitive decisions. Overall, with the exception of the (100\%, 50\%) case, all other pruning strategies outperform the baseline. This highlights the benefit of judiciously pruning spatial and temporal features, with our approach consistently improving model efficiency and performance across various pruning combinations.

\subsubsection{Analysis in Sparse Reward Settings}
\label{sec4}
In sparse reward environments, traditional reinforcement learning methods face significant challenges in credit assignment due to the delayed nature of rewards and the need to track which past actions contributed to success. Our proposed model addresses this by implementing a feature crediting mechanism that quantifies the importance of latent features at each decision step, enabling more precise credit assignment without heavy reliance on memory. The results in Table \ref{tab:performance_comparison_sparse} demonstrate the effectiveness of this approach across Medium and Medium-Expert environments. Our model consistently outperforms baselines, such as Decision Transformer (DT) and EDT, particularly in environments like HalfCheetah, Hopper, and Walker, where our method excels in sparse reward settings. For example, in the Medium-Expert HalfCheetah environment, our model achieves 81 ± 1.6, significantly surpassing the baseline (67 ± 1.2). Similarly, strong performance is observed in Walker and Hopper environments, demonstrating the robustness of our crediting mechanism across different tasks. This consistent improvement highlights our model's ability to efficiently learn from sparse feedback by focusing on relevant features, providing a more reliable solution for environments with infrequent or delayed rewards.

\section{Conclusion}

In this paper, we introduced AdaCred, a novel framework that addresses the challenges of suboptimal sequence representations in reinforcement learning through adaptive credit assignment and pruning of low-importance latent representations. By leveraging a causal latent graph structure, our approach efficiently identifies and retains critical components of the trajectory, improving both decision-making and computational efficiency. Theoretical analysis and empirical results on standard benchmarks demonstrate the effectiveness of AdaCred, surpassing state-of-the-art methods in both offline and imitation learning settings.

%%%%%%%%%%%%%%%%%%%%%%%%%%%%%%%%%%%%%%%%%%%%%%%%%%%%%%%%%%%%%%%%%%%%%%%%

%%% The acknowledgments section is defined using the "acks" environment
%%% (rather than an unnumbered section). The use of this environment 
%%% ensures the proper identification of the section in the article 
%%% metadata as well as the consistent spelling of the heading.

\begin{acks}
This work is based in part on research supported by SRC JUMP2.0 (CogniSense Center, Grant:
2023-JU-3133). The views, opinions,
findings, and conclusions expressed in this material are those of the author(s) and do not necessarily
reflect the official policies or positions of SRC.
\end{acks}

%%%%%%%%%%%%%%%%%%%%%%%%%%%%%%%%%%%%%%%%%%%%%%%%%%%%%%%%%%%%%%%%%%%%%%%%

%%% The next two lines define, first, the bibliography style to be 
%%% applied, and, second, the bibliography file to be used.

\bibliographystyle{ACM-Reference-Format} 
\bibliography{sample}

\appendix

\input{supp}

\end{document}

%% file: supp.tex
\startcontents[sections]
\printcontents[sections]{}{}{\setcounter{tocdepth}{3}} % Set 
% \appendix

%%%%%%%%%%%%%%%%%%%%%%%%%%%%%%%%%%%%%%%%%%%%%%%%%%%%%%%%%%%%%%%%%%%%%%%%

\appendix

\section*{\Large Appendix}

\begin{table*}
\centering
\adjustbox{max width=\textwidth}{
\begin{tabular}{@{}ll@{}}
\toprule
\textbf{Hyper-parameter} & \textbf{Value} \\ \midrule
Input sequence length ($T$) & 10, 20, 30 \\
Spatial Token Percentage & 75\% \\
Temporal Token Percentage & 75\% \\ 
 image size & 84 × 84, gray \\
Frame stack & 4 \\
Frame skip & 2 \\
Layers & 6 \\
MSA heads (Spatial Transformer)  & 8 \\
Embedding dimension (Spatial Transformer) & 192 \\
Patch size & 7 \\
MSA heads (Temporal Causal Transformer) & 4 \\
Embedding dimension (Temporal Causal Transformer) & 64 \\
Activation & GeLU  \\
Dropout & 0.1 \\
Learning rate & $6 \times 10^{-4}$ \\
Adam betas & (0.9, 0.95) \\
Grad norm clip & 1.0 \\
Weight decay & 0.1 \\
Learning rate decay & Linear warmup and cosine decay (see [9]) \\
Warmup tokens & $512 \times 20$ \\
Final tokens & $2 \times 500000 \times T$ \\ \bottomrule
\end{tabular}
}
\caption{Hyper-parameters used in the experiments for Atari }
\label{atari_hy}
\end{table*}

\section{Experimental Details}

\subsection{Baselines}

\textbf{Decision Transformer (DT) \cite{chen2021decision}:} The Decision Transformer (DT) reframes reinforcement learning as a sequence modeling problem, where the goal is to predict the next action given a history of past states, actions, and return-to-go values (i.e., the cumulative future rewards). DT uses a Transformer architecture, which is well-suited for capturing long-range dependencies in sequential data. Instead of traditional value-based approaches like Q-learning, DT treats RL as a supervised learning problem, training the model to output the best action based on previous states and rewards. The key innovation of DT is its ability to model sequences effectively while retaining information from long trajectories, making it a strong baseline in offline RL tasks where interaction with the environment is limited. However, DT may struggle in highly stochastic environments where predicting the return-to-go accurately is difficult, and its reliance on fixed-length trajectories can limit flexibility.

\textbf{Elastic Decision Transformer (EDT)\cite{wu2024elastic} :} The Elastic Decision Transformer (EDT) is an extension of the Decision Transformer that handles variable-length trajectories. Traditional Decision Transformers use fixed-length trajectory sequences, but in real-world scenarios, the length of trajectories can vary significantly. EDT introduces elastic mechanisms to handle this variability by using dynamic padding, masking, and positional encodings. This enables the Transformer to process trajectories of different lengths efficiently while maintaining the temporal order of the sequence. EDT also introduces adaptive mechanisms to prioritize certain trajectory segments based on their relevance to the final action prediction. EDT’s ability to adjust dynamically to different sequence lengths improves its flexibility in modeling trajectories of varying durations. While the model offers increased flexibility and performance on variable-length trajectories, the inclusion of elastic mechanisms also introduces additional complexity and computational costs, particularly with very short or long trajectories.

\textbf{Q-learning Decision Transformer (qDT) \cite{qlearning}:} The Q-learning Decision Transformer (qDT) combines the strengths of Q-learning with the sequence modeling capabilities of the Decision Transformer. qDT leverages dynamic programming principles, as used in Q-learning, to optimize the sequence modeling process. It conditions the prediction of actions not only on past states and rewards but also on dynamically calculated Q-values. By integrating Q-learning into the Transformer framework, qDT improves the model’s ability to handle suboptimal trajectories in offline RL tasks. This method is particularly useful in scenarios where trajectories include many suboptimal actions and rewards, as it helps guide the sequence model toward more optimal policy learning. The integration of Q-values adds robustness to the model's decision-making process, but also increases the computational complexity, especially in environments with large state-action spaces.

\textbf{CQL (Conservative Q-Learning) \cite{kumar2020conservative}:} Conservative Q-Learning (CQL) is an offline RL algorithm designed to prevent overestimation of Q-values, a common problem in offline settings where access to new data is restricted. CQL achieves this by penalizing overestimated Q-values, ensuring that the learned policy remains conservative and close to the behavior policy observed in the offline dataset. This conservativeness helps the algorithm avoid overfitting to over-optimistic actions, particularly when learning from suboptimal trajectories. CQL's focus on maintaining conservative estimates makes it highly effective in environments where distributional shifts between training and testing data are common, though it may underperform in settings that require more exploratory policies.

\textbf{QR-DQN (Quantile Regression Deep Q-Network)\cite{qrdqn2019}:} QR-DQN is a distributional variant of the Deep Q-Network (DQN) that estimates the entire distribution of future rewards rather than a single expected value. By predicting multiple quantiles of the reward distribution, QR-DQN captures the variability and uncertainty inherent in stochastic environments, allowing it to model not just the expected reward but also the risk associated with different actions. This ability to predict multiple quantiles allows QR-DQN to make more informed decisions in uncertain settings. However, the added complexity of estimating the entire reward distribution comes with increased computational demands, especially as the number of quantiles grows.

\textbf{REM (Random Ensemble Mixture) \cite{agarwal2020optimistic}:} REM (Random Ensemble Mixture) is an offline RL algorithm that utilizes an ensemble of Q-networks to improve the stability and robustness of Q-value estimates. By averaging the predictions of multiple Q-networks with random initializations, REM reduces the risk of overfitting and mitigates the inherent variance in Q-value estimates. This ensemble approach ensures that the learned policy is more robust to noisy or suboptimal data, making it highly suited for offline RL settings. The downside of this method is that training and maintaining multiple Q-networks can be computationally expensive, particularly as the number of Q-networks in the ensemble increases.

\textbf{BEAR (Bootstrapping Error Accumulation Reduction) \cite{kumar2019stabilizing}:} BEAR is another offline RL algorithm that focuses on reducing the accumulation of bootstrapping errors when learning from offline datasets. It does this by constraining the learned policy to remain close to the behavior policy observed in the offline data, using a regularization technique based on Maximum Mean Discrepancy (MMD). This prevents the learned policy from diverging too far from known actions, reducing the risk of compounding errors. While BEAR's ability to mitigate bootstrapping errors makes it a strong performer in offline settings, its heavy reliance on behavior policy constraints can limit the exploration of more optimal actions.

\subsection{Model Hyperparameters}
The detailed hyper-parameter settings for the Atari and GYM environments are provided in Tables \ref{atari_hy} and \ref{atari_hy}, respectively. Most hyper-parameters are aligned with those used in the Decision Transformer\cite{chen2021decision} and Starformer \cite{shang2022starformer}.

\begin{table*}
\centering
\adjustbox{max width=\textwidth}{
\begin{tabular}{@{}ll@{}}
\toprule
\textbf{Hyper-parameter} & \textbf{Value} \\ \midrule

Sequence length ($T$) & 20 \\
Image size & 84 × 84 \\
Spatial Token Percentage & 75\% \\
Temporal Token Percentage & 75\% \\ 
Frame stack & 3 \\
Frame skip & 4  \\
Layers & 3 \\
Activation & GeLU  \\
Dropout & 0.1 \\
Return-to-go conditioning & 6000 (HalfCheetah) \\
& 3600 (Hopper) \\
& 5000 (Walker) \\
& 50 (Reacher) \\
Learning rate & $1 \times 10^{-4}$ \\
Adam betas & (0.9, 0.95) \\
Grad norm clip & 1.0 \\
Weight decay & 0.1 \\
Learning rate decay & Linear warmup and cosine decay \\
Warmup tokens & $512 \times 20$ \\
Final tokens & $2 \times 100000 \times T$ \\ 
 \bottomrule
\end{tabular}
}
\caption{Hyper-parameters used in the experiments for gym}
\label{gym_hy}
\end{table*}

\section{Theoretical Discussion}

\subsection{Proof of Theorem 1: Minimal Sufficient Representations for Policy Learning}

\textbf{Theorem 1:}
\textit{Under the assumption that the causal graph \( G \) is Markov and faithful to the observed data, the set of minimal latent states \( \mathbf{g}_t^{\min} \subseteq \mathbf{g}_t \) is defined as:
\(
g_{i,t} \in \mathbf{g}_t^{\min} \iff c_{i}^{g \to r} = 1 \text{ or } g_{i,t} 
\) { has a directed path to a future reward through other latent states.}
This set forms a minimal and sufficient representation for policy learning, ensuring that only these latent state components are necessary to maximize the expected future rewards.}

We begin by establishing the minimal and sufficient representations for policy learning based on the Markov and faithfulness assumptions in the underlying causal graph \( G \).

\textbf{Definitions:}
\begin{itemize}
    \item \textbf{d-separation:} A path \( p \) in the causal graph \( G = (V, E) \) is said to be d-separated by a set of nodes \( Z \subseteq V \) if:
    \[
    X \perp_d Y \mid Z \quad \text{for any nodes } X, Y \in V.
    \]
    Specifically, \( Z \) d-separates \( X \) from \( Y \) if either:
    \begin{itemize}
        \item \( p \) contains a chain \( i \to m \to j \) or a fork \( i \gets m \to j \), where the middle node \( m \in Z \), or
        \item \( p \) contains a collider \( i \to m \gets j \) such that the middle node \( m \notin Z \) and no descendant of \( m \in Z \).
    \end{itemize}
    
    \item \textbf{Markov condition:} The Markov condition states that given the current latent state \( \mathbf{g}_t \), all future latent states \( \mathbf{g}_{t+k} \) are conditionally independent of past latent states \( \mathbf{g}_{t-k} \) and actions \( a_{t-k} \). Formally:
    \[
    P(\mathbf{g}_{t+k} \mid \mathbf{g}_t, a_t) = P(\mathbf{g}_{t+k} \mid \mathbf{g}_t, a_t, \mathbf{g}_{t-k}, a_{t-k}) \quad \text{for all} \ k > 0.
    \]
    
    \item \textbf{Faithfulness assumption:} The faithfulness assumption states that there are no conditional independencies between variables that are not entailed by the structure of the graph \( G \). That is, all independencies in the data are captured by the d-separation criterion of the graph.
\end{itemize}

\textbf{d-separation of irrelevant states}

Let \( \mathbf{g}_t = (g_{1,t}, g_{2,t}, \dots, g_{d,t})^\top \) denote the latent states at time \( t \). For any \( g_{i,t} \notin \mathbf{g}_t^{\min} \), we claim that it does not contribute to future rewards, i.e., it does not have a directed path to \( r_{t+k} \) for any \( k > 0 \). By the d-separation criterion, \( g_{i,t} \) is conditionally independent of future rewards, given the minimal latent states \( \mathbf{g}_t^{\min} \) and the cumulative reward \( R_{t+1} \):
\[
g_{i,t} \perp r_{t+k} \mid \mathbf{g}_t^{\min}, R_{t+1} \quad \text{for all} \ k > 0.
\]
Thus, pruning \( g_{i,t} \notin \mathbf{g}_t^{\min} \) does not affect the future rewards, and such states are irrelevant for policy learning.

\textbf{ Dependence of relevant states}

For any \( g_{i,t} \in \mathbf{g}_t^{\min} \), there exists a directed path from \( g_{i,t} \) to a future reward \( r_{t+k} \) for some \( k > 0 \). This implies that \( g_{i,t} \) contributes directly or indirectly to future rewards. Let \( \pi(g_{i,t}, r_{t+k}) \) denote the directed path from \( g_{i,t} \) to \( r_{t+k} \). 

By the Markov condition, \( g_{i,t} \) influences the future reward through direct or indirect interactions with other latent states in \( \mathbf{g}_t^{\min} \). Formally:
\[
P(r_{t+k} \mid \mathbf{g}_t^{\min}, a_t) = P(r_{t+k} \mid \mathbf{g}_t, a_t) \quad \text{for all } g_{i,t} \in \mathbf{g}_t^{\min}.
\]
If \( g_{i,t} \) were removed from \( \mathbf{g}_t^{\min} \), this would block a path to future rewards, violating the faithfulness assumption. Hence, every state \( g_{i,t} \in \mathbf{g}_t^{\min} \) is necessary for maximizing future rewards.

\textbf{ Minimality and Sufficiency}

We now show that \( \mathbf{g}_t^{\min} \) forms the minimal and sufficient set of latent states necessary for policy learning. By contradiction, assume that a state \( g_{i,t} \in \mathbf{g}_t^{\min} \) can be pruned. This would imply that the system could still predict future rewards without \( g_{i,t} \), which contradicts the Markov and faithfulness assumptions. 

Thus, the minimal sufficient set \( \mathbf{g}_t^{\min} \) is defined as:
\[
\mathbf{g}_t^{\min} = \{g_{i,t} \mid g_{i,t} \text{ contributes to } r_{t+k} \text{ for some } k > 0\}.
\]
By including all such \( g_{i,t} \) that influence future rewards, \( \mathbf{g}_t^{\min} \) guarantees that the agent can make optimal decisions to maximize the expected future rewards.

\qed

\subsection{Proof of Theorem 2: Structural Identifiability}

\textbf{Theorem 2:}
\textit{Suppose the underlying latent states \( \mathbf{g}_t \) are observed, i.e., the environment follows a Markov Decision Process (MDP). Then, under the Markov condition and faithfulness assumption, the structural matrices \( \mathbf{C}^{g \to g}, \mathbf{C}^{a \to g}, \mathbf{C}^{g \to r}, \mathbf{C}^{a \to r} \) are identifiable.}

We now rigorously prove that the structural matrices \( \mathbf{C}^{g \to g}, \mathbf{C}^{a \to g},\)
\( \mathbf{C}^{g \to r}, \mathbf{C}^{a \to r} \), which encode the causal relationships between latent states, actions, and rewards, are identifiable under the Markov condition and faithfulness assumption.

\textbf{Global Markov Condition and Faithfulness}

Let \( G = (V, E) \) be a directed acyclic graph (DAG) representing the dynamic Bayesian network (DBN) for the latent states \( \mathbf{g}_t \), actions \( a_t \), and rewards \( r_t \). The Global Markov Condition implies that for any disjoint sets of nodes \( X, Y, Z \subseteq V \), if \( Z \) d-separates \( X \) from \( Y \), then \( X \) is conditionally independent of \( Y \) given \( Z \). Formally:
\[
X \perp Y \mid Z \quad \text{if and only if } Z \text{ d-separates } X \text{ and } Y \text{ in } G.
\]
The structure of the DBN assumes that the temporal evolution of the environment can be represented by the following conditional distributions:
\[
P(\mathbf{g}_{t+1} \mid \mathbf{g}_t, a_t) = P(\mathbf{g}_{t+1} \mid \mathbf{g}_t, a_t),
\]
\[
P(r_t \mid \mathbf{g}_{t-1}, a_t) = P(r_t \mid \mathbf{g}_{t-1}, a_t),
\]
where the causal relationships between variables are encoded by the structural matrices \( \mathbf{C} \).

The faithfulness assumption ensures that any observed conditional independencies in the data reflect the corresponding d-separations in the graph \( G \). Therefore, the structural relationships encoded in \( \mathbf{C}^{g \to g}, \mathbf{C}^{a \to g}, \mathbf{C}^{g \to r}, \mathbf{C}^{a \to r} \) must be identifiable from the observed data.

\textbf{Identifiability of Structural Matrices}

We now show that each structural matrix \( \mathbf{C} \) is identifiable by leveraging the conditional dependencies between variables.

\begin{itemize}
    \item \textbf{Identifiability of \( \mathbf{C}^{g \to g} \)}: The matrix \( \mathbf{C}^{g \to g} \in \{0,1\}^{d \times d} \) encodes the causal influence of the previous latent states \( \mathbf{g}_{t-1} \) on the current latent states \( \mathbf{g}_t \). For each entry \( c_{i,j}^{g \to g} \), we have:
    \[
    c_{i,j}^{g \to g} = 1 \iff \mathbf{g}_{i,t} \not\perp \mathbf{g}_{j,t-1} \mid \mathbf{g}_{\setminus \{i\}, t-1}, a_{t-1},
    \]
    meaning \( g_{i,t} \) is conditionally dependent on \( g_{j,t-1} \), which is identifiable by observing conditional dependencies in the data. Conversely, if \( c_{i,j}^{g \to g} = 0 \), the variables are conditionally independent, and this independence is similarly identifiable through d-separation in the graph \( G \).
    
    \item \textbf{Identifiability of \( \mathbf{C}^{a \to g} \)}: The matrix \( \mathbf{C}^{a \to g} \in \{0,1\}^{d} \) encodes the causal effect of actions \( a_t \) on the latent states \( \mathbf{g}_t \). For each entry \( c_i^{a \to g} \), we have:
    \[
    c_i^{a \to g} = 1 \iff g_{i,t} \not\perp a_t \mid \mathbf{g}_{t-1},
    \]
    which implies that \( a_t \) has a direct causal effect on \( g_{i,t} \). This causal dependency can be identified by analyzing the conditional dependence between \( g_{i,t} \) and \( a_t \). If \( c_i^{a \to g} = 0 \), then \( g_{i,t} \) is conditionally independent of \( a_t \), which can be identified by observing the corresponding d-separations in the data.
    
    \item \textbf{Identifiability of \( \mathbf{C}^{g \to r} \)}: The matrix \( \mathbf{C}^{g \to r} \in \{0,1\}^{d} \) encodes the causal influence of the latent states \( \mathbf{g}_{t-1} \) on the reward \( r_t \). For each entry \( c_i^{g \to r} \), we have:
    \[
    c_i^{g \to r} = 1 \iff r_t \not\perp g_{i,t-1} \mid \mathbf{g}_{\setminus \{i\}, t-1}, a_{t-1},
    \]
    which implies that \( r_t \) is conditionally dependent on \( g_{i,t-1} \). This can be identified by analyzing the conditional dependence structure of \( r_t \) and the latent state variables \( g_{i,t-1} \). If \( c_i^{g \to r} = 0 \), the variables are conditionally independent, and this independence is identifiable through d-separation.
    
    \item \textbf{Identifiability of \( \mathbf{C}^{a \to r} \)}: The matrix \( \mathbf{C}^{a \to r} \in \{0,1\} \) encodes the causal influence of actions \( a_t \) on the reward \( r_t \). For the entry \( c^{a \to r} \), we have:
    \[
    c^{a \to r} = 1 \iff r_t \not\perp a_t \mid \mathbf{g}_{t-1},
    \]
    which implies that \( a_t \) directly affects \( r_t \). This causal dependence is identifiable from the data. If \( c^{a \to r} = 0 \), then \( r_t \) and \( a_t \) are conditionally independent, and this independence can be determined by observing the lack of conditional dependencies in the data.
\end{itemize}

By showing that the structural matrices \( \mathbf{C}^{g \to g}, \mathbf{C}^{a \to g}, \mathbf{C}^{g \to r}, \mathbf{C}^{a \to r} \) encode relationships that are identifiable through conditional dependencies and independencies in the data, we conclude that these matrices are identifiable under the Markov condition and faithfulness assumption.

\textbf{Applications to Online RL and Computer Vision:} While this work focuses on offline RL, the proposed method has potential applications in a broader context. It can be extended to transformer-based online RL \cite{kumawat2024robokoop}, as well as computer vision tasks such as detection and classification \cite{9448387, 9892184, 9892390, 10600387}. Furthermore, it can be applied to multi-agent modeling using transformer networks \cite{kumawat2024stemfold, kumawat2024stage}.

\qed